# Hippocampus-DETR: An Explicit Memory Object Detection Framework Based on Hippocampus Modeling

Zhaoning Shi (1), Bo Ma (1), Hao Xu (1), Zepeng Yang (1), Bo Liang (1)
((1) Beijing institute of Technology)

**Abstract**

This paper addresses the lack of explicit memory mechanisms in current object detection models and proposes Hippocampus-DETR, a novel detection framework based on biological hippocampal memory modeling. This framework integrates a hippocampal memory network module, HipNet, into the DETR architecture and systematically simulates the anatomical structure and functional organization of hippocampal subregions, including the entorhinal cortex, dentate gyrus, CA3, CA1, and subiculum. Through this design, Hippocampus-DETR realizes pattern separation, pattern completion, importance filtering, and information integration of visual encoding features. During training, different memory submodules are optimized using a layer-wise training strategy, ultimately forming a memory system with memory retrieval and completion capabilities. Experimental results demonstrate that Hippocampus-DETR achieves higher detection accuracy than current mainstream models. More importantly, models equipped with this framework also exhibit excellent generalization ability and data efficiency in tasks such as few-shot image classification, multimodal feature construction, and image restoration. Subsequent experiments further validate the functional necessity and internal interpretability of each memory submodule. This study not only provides a novel object detection framework, but also offers a feasible technical pathway for integrating neurocognitive mechanisms with deep learning models, highlighting its significant value in improving model learning efficiency and task robustness. The project is available at https://github.com/2186cloud/hipnet.

## 1 Introduction

Object detection, as one of the core tasks in computer vision, plays a crucial role in fields such as autonomous driving and intelligent robotics, and has achieved remarkable progress in recent years with the rapid development of deep learning. The DETR [1] model is a mainstream method in object detection and has long been extensively studied and applied. However, with the introduction of more complex methods and the continuous expansion of model scale, the difficulty of model learning has gradually increased, thereby limiting further improvements in model performance.

To improve the performance of the original DETR model, various enhanced variants have subsequently been proposed. Deformable DETR [2] introduces deformable attention, which attends to a small set of key sampling points around reference points, thereby alleviating the problems of high computational cost and slow convergence, while incorporating multi-scale features to enhance perception. To simplify the architecture, Efficient DETR [3] innovatively combines dense and sparse detection by using an RPN to generate proposals for initializing object queries, thus substantially reducing the decoder to a single layer and improving training efficiency. To address computational redundancy in the encoder, Sparse DETR [4] proposes token sparsification, which uses objectness scores or decoder cross-attention maps (DAM) to select informative tokens for computation, reducing encoder complexity. For real-time scenarios, RT-DETR [5] designs a hybrid encoder for multi-scale feature fusion and adopts an IoU-aware query selection strategy to initialize queries. Together with a lightweight decoder, this achieves a balance between speed and accuracy. DINO [6] significantly improves training stability and final performance by introducing advanced training techniques such as denoising training and contrastive denoising. These variants have collectively advanced the development and practical application of the DETR framework from different perspectives, including attention mechanisms, architectural design, computational efficiency, and training strategies.

Overall, although the aforementioned improvements to DETR enhance model efficiency and accuracy from different perspectives, they share a fundamental architectural limitation compared with human-like learning: the absence of an explicit memory mechanism. This limitation leads to further technical bottlenecks, such as difficulty in handling occlusion. When an object is severely occluded, the lack of a mechanism for memorizing its previous complete state may result in missed detections. Memory capability enables the model to explicitly complete partial features and subsequently use the recalled complete features for downstream detection tasks, thereby achieving more robust detection performance. In addition, memory can allow the model to recall learned features from less familiar inputs, improving learning efficiency. In summary, incorporating memory capability into object detection models is an effective way to further improve object detection performance.

In this work, we propose Hippocampus-DETR, which aims to introduce the memory function of the hippocampus into object detection models. Unlike studies that are merely inspired by memory, we construct a hippocampal neural network module, termed HipNet, from both anatomical and functional perspectives, and integrate it into the DETR architecture. Anatomically, HipNet incorporates key hippocampal subregions, including the dentate gyrus (DG), CA3, CA1, subiculum, and entorhinal cortex(EC), as well as the major projection pathways among them. Functionally, it models sensory transmission in the EC, pattern separation in the DG, pattern completion in CA3, output comparison in CA1, and information integration in the subiculum. Experimental results show that Hippocampus-DETR achieves higher detection accuracy than current mainstream models. Further validation demonstrates that it can reduce data dependence and performs well in few-shot image classification. It also exhibits unique advantages and potential in other AI tasks, such as multimodal learning, while maintaining good internal interpretability. Ablation experiments on individual subregions further confirm the effectiveness of the functional differentiation among these modules.

Overall, the key contributions of this paper are summarized as follows:

- We propose a key insight that introducing memory functionality into object detection and image classification models can improve the robustness and learning efficiency of models.
- To the best of our knowledge, this is the first work to integrate a hippocampal memory module constructed based on biological anatomical structures and functions into the DETR architecture, thereby improving its object detection performance.
- Extensive experiments not only demonstrate the effectiveness of our method in object

detection, but also reveal the advantages of introducing memory functionality in broader domains, such as few-shot image classification.

## 2 Related work

### 2.1 Object detection model

Currently, there are two mainstream branches in object detection: one-stage regression models represented by YOLO and end-to-end detection Transformer models represented by DETR. The YOLO model [7] first reformulated object detection as a single regression problem, significantly improving detection speed and accuracy. Subsequent iterations have continuously optimized its architecture and training strategies. From YOLOv4 to YOLOv8 [8-12], the YOLO series gradually incorporated modules such as CSP, SPP, PANet, multimodal support, and anchor-free detection heads, achieving a better balance between throughput and accuracy. YOLOv9 [13] and YOLOv10 [14] further focused on lightweight backbone networks and simplified deployment. YOLOv11 [15] replaces the original C2f module with a more efficient C3k2 unit and introduces convolutional blocks with local spatial attention, enhancing its ability to detect small and concealed objects. The latest YOLOv12 [16] introduces residual efficient layer aggregation networks, lightweight area attention, and flash attention mechanisms, enabling efficient global and local semantic modeling while maintaining real-time performance, thereby further improving robustness and accuracy. YOLOv13 [17] proposes a hypergraph-based adaptive correlation enhancement mechanism, HyperACE, which adaptively exploits latent higher-order correlations. This overcomes the limitation of previous methods that rely only on pairwise correlation modeling and enables efficient cross-location and cross-scale feature fusion and enhancement. The DETR series achieves truly end-to-end object detection by eliminating non-maximum suppression (NMS). However, its early versions suffer from slow training convergence, high computational cost, and difficulty in query optimization. To address these challenges, researchers have proposed various improved architectures. Deformable DETR improves training efficiency under multi-scale features through deformable attention. DAB-DETR [18] and DN-DETR [19] introduce iterative refinement and denoising training strategies, respectively, to further improve detection performance. Group-DETR [20] adopts a grouped one-to-many assignment strategy to reduce computational overhead. Efficient DETR and Sparse DETR reduce complexity by decreasing the number of encoder-decoder layers or limiting the number of query updates. Lite DETR [21] improves encoder efficiency through an interleaved update mechanism that reduces the update frequency of shallow features. Conditional DETR [22] and Anchor DETR [23] reduce optimization difficulty by improving query design. DINO proposes a mixed query selection mechanism to improve query initialization. RT-DETR further explores improvements in computational efficiency and query initialization strategies. In summary, most existing improvements in object detection models focus on enhancing feature association capability. Building upon this line of research, this paper investigates the introduction of a memory mechanism to further improve the performance of DETR-based models.

### 2.2 Object Detection with Memory

Several object detection algorithms that incorporate memory have been proposed. For example, [24] introduces a retrieval-augmented classification (RAC) module and a memory bank that can be flexibly updated with knowledge from new domains, enabling the detector to retrieve similar object concepts from memory during testing. MAD [25] effectively aligns and fuses object proposals from the detector with those from a memory bank of past predictions, improving the performance of LiDAR- and camera-based 3D object detection networks. YOLO LwF [26], when combined with replay memory, can significantly mitigate forgetting and improve object detection performance. In addition, Titans proposes a novel neural long-term memory module that can learn to memorize historical context and assist the attention mechanism in better focusing on the current context while utilizing long-term historical information [27]. These studies validate the importance of memory for object detection from multiple perspectives. However, although they construct various memory storage modules inspired by human memory, they do not systematically investigate how memory is implemented in the brain. In contrast, we conduct a detailed study and modeling of hippocampal subregions and their projection relationships, thereby constructing a more effective memory network architecture. We further apply it to RT-DETR, resulting in an object detector with higher accuracy.

## 3 Hippocampal Formation

Since our memory module primarily emulates the hippocampal formation, it is necessary to first clarify the scope of the subregions involved in the structure we model, as well as their major projection relationships, before introducing our model.

Our model emulates the hippocampal formation by incorporating the DG, CA3, CA1, subiculum, and EC [28]. These regions are generally regarded as the most fundamental areas involved in hippocampal memory function [29-31]. The EC is typically considered to have a six-layer structure (EC1-EC6) [32]. Although EC1 may play a role in feedforward inhibition, it contains only a small number of interneurons and primarily projects to layers II and III of the EC [32,33]. This likely suggests that EC1 is not an independent information-processing structure, but rather assists the functions of EC2 and EC3. Therefore, modeling EC2 and EC3 alone is sufficient to implement the corresponding functions, without explicitly modeling EC1. This simplification does not affect the subsequent projection relationships among hippocampal subregions. EC4 also contains relatively few neurons, and its cellular morphology and physiological properties are similar to those of EC3 and EC5 [32]. Meanwhile, EC3 has extensive projections to EC5 [34], indicating that assigning EC4 neurons to EC3 and EC5 would have only a minor impact on the functions and projections of the EC. Therefore, EC4 is not modeled separately. EC5 and EC6 are generally regarded as output layers that integrate information from CA1 and the subiculum and project to regions such as forebrain structures [35]. Since this paper does not investigate these downstream projection targets, EC5 and EC6 are treated as a unified structure in our modeling and are referred to as EC5 for simplicity.

For these subregions, the anatomical connectivity of the hippocampal formation emulated by HipNet is as follows. Sensory information, although multiple regions within the hippocampal formation receive projections from other brain areas [32], is considered here primarily in terms of sensory input, which mainly enters the hippocampal formation through EC2 [36]. Other external inputs to the hippocampal formation are ignored in our model. In addition to projecting to EC3, EC2 mainly projects to the DG and CA3 via the perforant pathway [37]. The DG also projects to CA3 through mossy fibers. CA3 primarily projects to CA1 and to itself. In addition to projecting to EC5, as a result of merging EC4 into EC3 and EC5, the projections involving EC4 are also incorporated into this pathway, EC3 also projects to CA1 and the subiculum via the temporoammonic pathway. CA1 subsequently projects mainly to the subiculum and EC5. The subiculum ultimately projects to EC5, with the EC4 region receiving subicular projections fully incorporated into EC5 in our model [38]. Finally, both the subiculum and EC5 project outward [39]. Although other projection relationships also exist within the hippocampal formation [40] and may influence hippocampal function to varying degrees, HipNet is designed as a foundational framework connecting neuroscience and deep learning. Therefore, while maintaining a relatively complete structure, we aim to minimize less important connections in HipNet. Accordingly, we retain only the unidirectional projections among subregions that are important for information transmission in subsequent modeling.

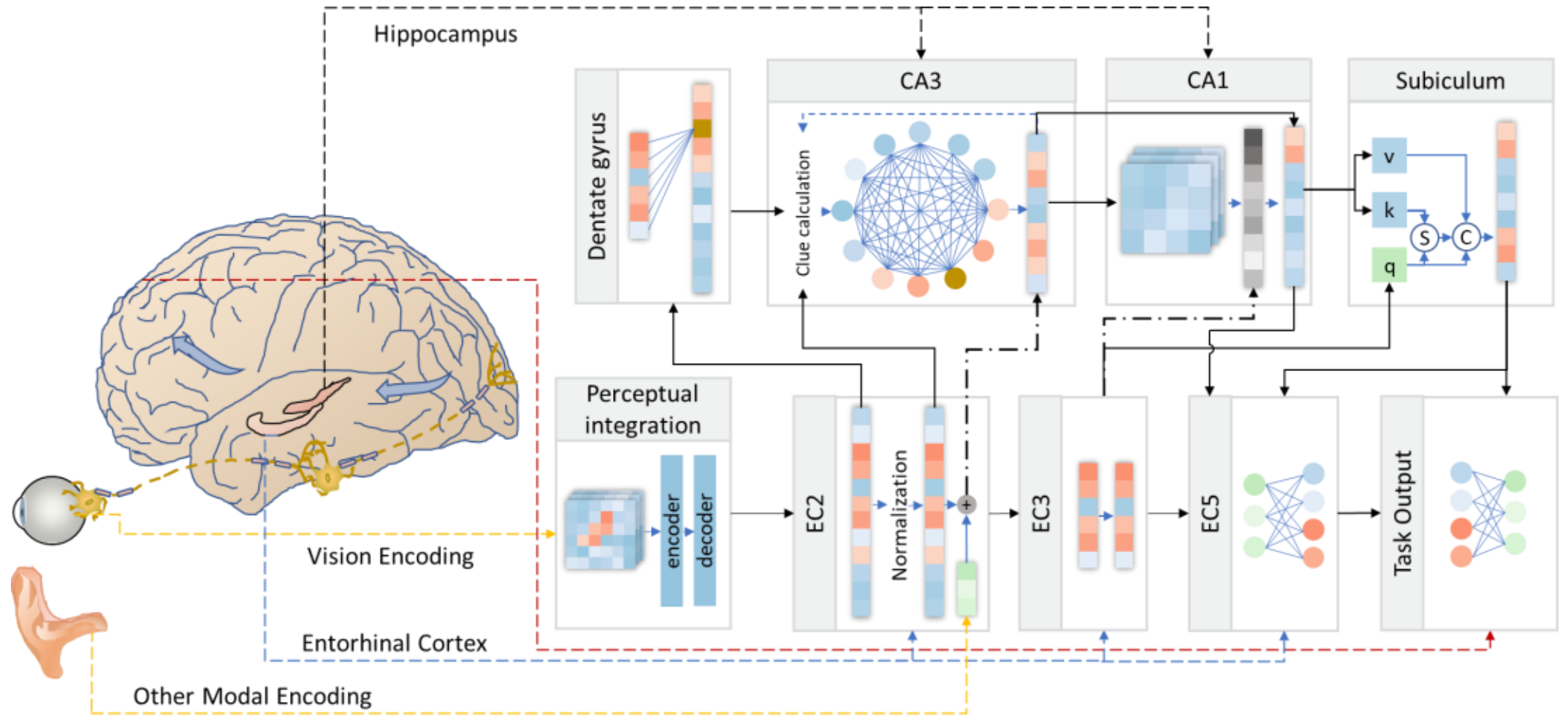

**Fig.1| The overall structure of HipNet.** Each gray box represents a modeled subregion module. Solid arrows indicate actual projection relationships. Dash-dotted lines represent information flows that only participate in model optimization during the training phase. Dashed arrows represent the correspondence between biological concepts and the model.

## 4 Method

While emulating the hippocampal formation, the hippocampal memory model must also account for the requirements of practical tasks. Here, we mainly design the model for object detection. As shown in Fig. 1, the overall model can be divided into three components: the perceptual integration component, the memory component (EC2, EC3, EC5, DG, CA3, CA1, and SUB), and the detection output component. This work focuses on exploring the construction of the memory component corresponding to the hippocampus, while the perceptual integration and detection output components are designed based on RT-DETR-R50. The most important ideas and module designs are described below. More detailed descriptions of each component and the training settings are provided in the supplementary methods.

**Perceptual Integration and Detection Output Components.** The perceptual integration component provides effective features for the memory component, playing a role analogous to the eyes and hierarchical visual cortices. We use a pretrained RT-DETR as the network model for this component, where the final output tokens of its decoder serve as the output information transmitted from the perceptual integration component to the memory component. Only tokens whose corresponding class confidence exceeds a threshold of 0.9 are regarded as valid output information; otherwise, they are ignored. The detection output component follows the same operations as RT-DETR, except that the original decoder embeddings of RT-DETR are replaced with the embeddings output by the memory component.

**EC.** From the perspective of its role in information processing, the EC participates in the interconnection between the hippocampal formation and the cortex [41,42], primarily serving as a hub for information transmission. Accordingly, in our model, the EC is mainly responsible for receiving features from the perceptual integration component and transmitting them to the DG, CA3, CA1, the subiculum, and the subsequent detection output component. EC2 includes normalization and empty-feature elimination operations, which are crucial for enabling the functions of the subsequent modules. EC3 only serves to transmit features to the corresponding modules. EC5 uses a fully connected neural network to fuse feature information from EC3, CA1, and the subiculum, and then projects it outward.

**DG.** The DG is considered to support pattern separation, generating distinct sparse representations for highly similar events [43,44]. Such sparse representations facilitate the subsequent pattern completion function of CA3 [45,46]. However, some studies have questioned the importance of DG projections to CA3, suggesting that the DG may mainly play a modulatory role [47]. Although the function of the DG has not yet been fully understood in neuroscience, CA3 primarily receives projections from the DG and EC2 within the hippocampal formation. It is therefore reasonable to assume that the DG and EC2 jointly transmit information to CA3 to support its pattern completion function. Based on these facts and reasoning, the function of the DG in the memory component is designed to generate sparse representations that are as orthogonal as possible for different features. These representations cooperate with the output features of EC2 to ultimately produce cue features that support pattern completion in CA3. In the memory component, the DG uses a self-organizing map (SOM) network to perform self-organized learning on features from EC2 [48], ensuring representational sparsity through a small number of winning neurons. Subsequently, the memorized EC2 output features are treated as the kernels of fuzzy sets, and the cosine similarity between the current EC2 output feature and each memorized feature is used as the membership function. The membership degree of the current EC2 output feature to these fuzzy sets is then calculated, thereby mapping the current feature to similar memorized features. Finally, the kernels of the fuzzy sets corresponding to each feature are merged and provided to CA3 as cue features. Since the features corresponding to these kernels have already been designed to participate in CA3 training during the training stage, reactivating these features is beneficial for pattern completion in CA3.

**CA3.** CA3 is widely regarded as the principal hippocampal region responsible for pattern completion, meaning that it can trigger a comprehensive recall of a memorized item from a cue [49]. Similarly, in the memory component, CA3 is designed as an auto-associator that uses input cues to output the corresponding memory information, thereby achieving pattern completion. Structurally, CA3 exhibits extensive CA3-to-CA3 connectivity; therefore, CA3 is often modeled as a recurrent neural network to simulate its structure and function. Although models that are more biologically plausible, such as continuous attractor neural networks, are often used in computational neuroscience to simulate the pattern completion function of CA3 [50], here we employ a modern Hopfield network [51], which is more suitable for AI tasks, to model CA3. First, the Hopfield network shares a certain degree of similarity with CA3. On the one hand, the output information of a Hopfield network is fed back as input, which structurally corresponds to the internal CA3-to-CA3 projections. On the other hand, Hopfield networks are commonly used to retrieve and store patterns, which is analogous to the function of pattern completion. In addition, the modern Hopfield network used in this paper has been applied as a deep learning component in practical applications and has demonstrated its effectiveness [51]. Therefore, we choose it as the model for the CA3 region in the memory component. Here, the Hopfield network takes the cue features computed by DG and EC2 as input and outputs the memorized EC2 features stored during training. The EC2 features include not only convolution-derived visual features, but also one-hot encodings of labels. These one-hot encodings are regarded as encodings of the same object from other sensory modalities, such as textual word embeddings.

**CA1.** The CA1 region of the hippocampus is considered to support functions such as novelty detection, cross-layer coordination, and output comparison [52], all of which involve the coordinated processing of memory information and novel information. Meanwhile, considering that CA1 receives memory information from CA3 and current sensory information from EC3, we infer that CA1 may compare current sensory input with memory representations and thereby retrieve memory features that are more important for the current task. We implement this function using a convolutional neural network. The convolutional neural network used to model CA1 takes visual memory features from CA3 as input and outputs the importance of these features. For different tasks, the criteria for measuring feature importance may vary. Here, we assume a common image classification scenario. CA1 is trained using images different from those memorized by CA3. Along the channel dimension, we compare the current features output by EC3 with the memory features output by CA3 for the corresponding category using cosine similarity. When the similarity exceeds a specified threshold, the feature is considered, to some extent, a shared feature of objects in that category and will be retained in subsequent processing. Otherwise, the feature is regarded as instance-specific and will be ignored subsequently.

**SUB.** The subiculum connects the hippocampus with other brain regions and is generally believed to be responsible for integrating and distributing information processed by the hippocampal system [53,54]. It is also the major origin of the largest number of hippocampal efferent projections [55]. Since investigating the subsequent outward projections of the subiculum and EC5 is beyond the scope of this paper, we simplify the brain regions receiving projections from the subiculum and EC5 into a single entity, namely the detection output component. Under this premise, the difference between the subiculum and EC5 lies more in the content of their output information than in their subsequent projection relationships. Because the EC primarily transmits current sensory information, the subiculum in this paper is designed for the integration and output of memory information. Meanwhile, the subiculum receives projections from CA1 and EC3; that is, the subiculum here uses both memory information and current sensory information as inputs to output integrated memory information. This is highly conducive to producing post-event misinformation memory, in which subjects incorporate incorrect information provided after an event into their memory of the event. Such memory distortion has been considered by many studies to contribute to the effective functioning of cognitive systems [56]. Here, we design a "memory attention" mechanism to model the subiculum. Specifically, the current sensory features from EC3 are used as the query (q), while the memory features from CA1 are used as the key (k) and value (v). When the q corresponding to the current sensory features and the k corresponding to the memory features are judged to be similar, the subiculum outputs the memory features, indicating that the memory is not "distorted" at this time. When q and k are judged to be dissimilar, the subiculum outputs the current sensory features, indicating that the memory is "distorted" by sensory input. To enhance robustness, subiculum is trained using all detection boxes in the training set with confidence scores greater than 0.5.

**Training and Testing.** For the DG and CA3 modules, one sample from each class is used for training to obtain a memory prototype for each category. The CA1 module is trained using the remaining samples, excluding those used to train the DG and CA3 modules, to strengthen its ability to identify important features. All other modules are trained using the full training set. During testing, memory is used for auxiliary judgment only when the predicted category of the HipNet branch is inconsistent with that of RT-DETR. To ensure that the memory involved in prediction has sufficient confidence and that RT-DETR has relatively low confidence for the target, we apply the following criterion: when the maximum value in the 80-dimensional classification vector predicted by RT-DETR is less than 0.9 and the difference between the maximum and second-largest values is less than 0.1, indicating high category confusion, while the maximum value in the HipNet prediction vector is greater than 4.5 and the difference between the maximum and second-largest values is greater than 0.19, indicating low category confusion, the maximum value from HipNet is substituted into the corresponding position of the RT-DETR output. In this way, memory is used to improve model performance while minimally altering the original output. To ensure that the detection boxes contain objects during testing and that memory can improve accuracy, the above memory assistance is applied only to detection boxes with RT-DETR confidence scores between 0.5 and 0.9. In all other cases, the RT-DETR output is used directly.

## 5 Experimental Results

### 5.1 Settings

HipNet is a portable module. Therefore, in addition to applying Hippocampus-DETR to the object detection task, we further evaluate its effectiveness on multiple tasks, including few-shot image classification, multimodal learning, and image restoration. The results consistently show that introducing hippocampal memory capability brings unique advantages and broad research value. Subsequent ablation experiments demonstrate the effectiveness of modeling each hippocampal subregion. Finally, we present the local interpretability of the hippocampal memory module.

The number of training epochs, learning rate, and other parameters differ across modules. We provide detailed descriptions of these hyperparameter settings in the Supplementary Information. The batch size is set to 8. Training and testing are conducted on a single NVIDIA Tesla V100 GPU.

### 5.2 Comparative Results on Object Detection

Table I presents the comparative results on the MS COCO dataset [57]. The proposed method is compared with YOLO- based and DETR-based models of comparable scale. As shown in the table, our Hippocampus-DETR achieves state-of-the-art performance. More importantly, since HipNet is a module independent of RT-DETR, it can be broadly transferred to other models to improve their performance. Meanwhile, memory is more advantageous in tasks with limited training samples. On the COCO dataset, where the amount of data is large, this effect is less pronounced; we validate and discuss this point in Section 5.3. Even though the advantage of memory is diluted by large-scale data, Hippocampus-DETR still achieves improved accuracy compared with the original RT-DETR. This improvement is attributed to the more robust HipNet branch, which corrects certain misclassifications made by RT-DETR.

Table I

QUANTITATIVE COMPARISON WITH OTHER STATE-OF-THE-ART REAL-TIME OBJECT DETECTORS ON MS COCO DATASET.

| Method | FLOPs(G) | Parameters (M) | $AP^{val}_{50:95}$ | $AP^{val}_{50}$ | $AP^{val}_{75}$ |
|---|---|---|---|---|---|
| YOLOv6-3.0-N[10] | 11.4 | 4.7 | 37 | 52.7 | – |
| YOLOv6-3.0-S[10] | 45.3 | 18.5 | 44.3 | 61.2 | – |
| YOLOv6-3.0-L[10] | 150.7 | 59.6 | 51.8 | 69.2 | – |
| Gold-YOLO-N[58] | 12.1 | 5.6 | 39.6 | 55.7 | – |
| Gold-YOLO-S[58] | 46 | 21.5 | 45.4 | 62.5 | – |
| Gold-YOLO-L[58] | 151.7 | 75.1 | 51.8 | 68.9 | – |
| YOLOv8-N[12] | 8.7 | 3.2 | 37.4 | 52.6 | 40.5 |
| YOLOv8-S[12] | 28.6 | 11.2 | 45 | 61.8 | 48.7 |
| YOLOv8-L[12] | 165.2 | 43.7 | 53 | 69.8 | 57.7 |
| YOLOv9-S[13] | 26.4 | 7.1 | 46.8 | 63.4 | 50.7 |
| YOLOv9-C[13] | 102.1 | 25.3 | 53 | 70.2 | 57.8 |
| YOLOv10-N[14] | 6.7 | 2.3 | 38.5 | 53.8 | 41.7 |
| YOLOv10-S[14] | 21.6 | 7.2 | 46.3 | 63 | 50.4 |
| YOLOv10-L[14] | 120.3 | 24.4 | 53.2 | 70.1 | 57.2 |
| YOLO11-N[15] | 6.5 | 2.6 | 38.6 | 54.2 | 41.6 |
| YOLO11-S[15] | 21.5 | 9.4 | 45.8 | 62.6 | 49.8 |
| YOLO11-L[15] | 86.9 | 25.3 | 52.3 | 69.2 | 55.7 |
| YOLOv12-N[16] | 6.5 | 2.6 | 40.1 | 56 | 43.4 |
| YOLOv12-S[16] | 21.4 | 9.3 | 47.1 | 64.2 | 51 |
| YOLOv12-L[16] | 88.9 | 26.4 | 53 | 70 | 57.9 |
| YOLOv13-N[17] | 6.4 | 2.5 | 41.6 | 57.8 | 45.1 |
| YOLOv13-S[17] | 20.8 | 9 | 48 | 65.2 | 52 |
| RT-DETR-R18[5] | 60 | 20 | 46.4(46.4006) | 63.7(63.7204) | 50.3(50.2757) |
| RT-DETR-R50[5] | 136 | 42 | 53.1(53.0536) | 71.2(71.2180) | 57.7(57.6708) |
| LW-DETR-L[59] | 72 | 47 | 49.5 | – | – |
| LW-DETR-X[59] | 174 | 118 | 53 | – | – |
| Hippocampus-DETR-R18 | 75 | 47.1 | 46.4(46.4010± 0.0003) | 63.7(63.7209± 0.0003) | 50.3(50.2758± 0.0001) |
| Hippocampus-DETR-R50 | 153.0 | 69.4 | **53.1(53.0539± 0.0002)** | **71.2(71.2187± 0.0003)** | 57.7(57.6713± 0.0001) |

Note: The accuracy of RT-DETR-R50 is obtained by reproducing the results using the weights provided in the original paper.

### 5.3 Comparative Results on Few-Shot Image Classification Tasks

Next, to verify whether the hippocampal memory module can help reduce the data requirement of the model, we tested it on few-shot image classification tasks. To transfer HipNet to image classification models, we mainly adjusted its input and output interfaces. For the input part, each input image is first divided into nine smaller regions, and independent convolutional operations are then performed on each region to extract features, which are subsequently fed into the memory component. Compared with applying convolution to the whole image, regional convolution forces the input sources corresponding to the final output channels to be differentiated, thereby facilitating the orthogonalization of output features. For the output part, we directly use a two-layer fully connected network to integrate the features from EC5 and subiculum and predict the category. More details on the settings are provided in the supplementary methods. Since the hippocampal memory module uses a single sample as the memory prototype, it is naturally suitable for few-shot tasks. Meanwhile, conventional few-shot learning commonly relies on pretraining with a base dataset, which does not reduce the data volume or annotation requirement of the base dataset. Moreover, conventional experiments usually require the feature distributions of the base and novel datasets to be relatively consistent in order to achieve good performance. Therefore, we argue that directly training the model with few-shot samples, rather than using large-scale base data for pretraining, better demonstrates the improvement in data utilization efficiency brought by the memory module. In this paper, we train the model directly using a few-shot support set and then evaluate its classification performance on a large-scale query set to verify the learning ability of the hippocampal memory module. Experiments are conducted on the FashionMNIST, MNIST, and CIFAR-10 datasets [60-62]. For the DG and CA3 modules, only one sample per class is likewise used for training to obtain prototype memory for each category. The CA1 module is trained using the remaining samples, excluding those used to train the DG and CA3 modules. All other modules are trained using the entire few-shot training set. Since CA1 requires samples different from those used for training the DG and CA3 modules, the minimum number of shots is 2.

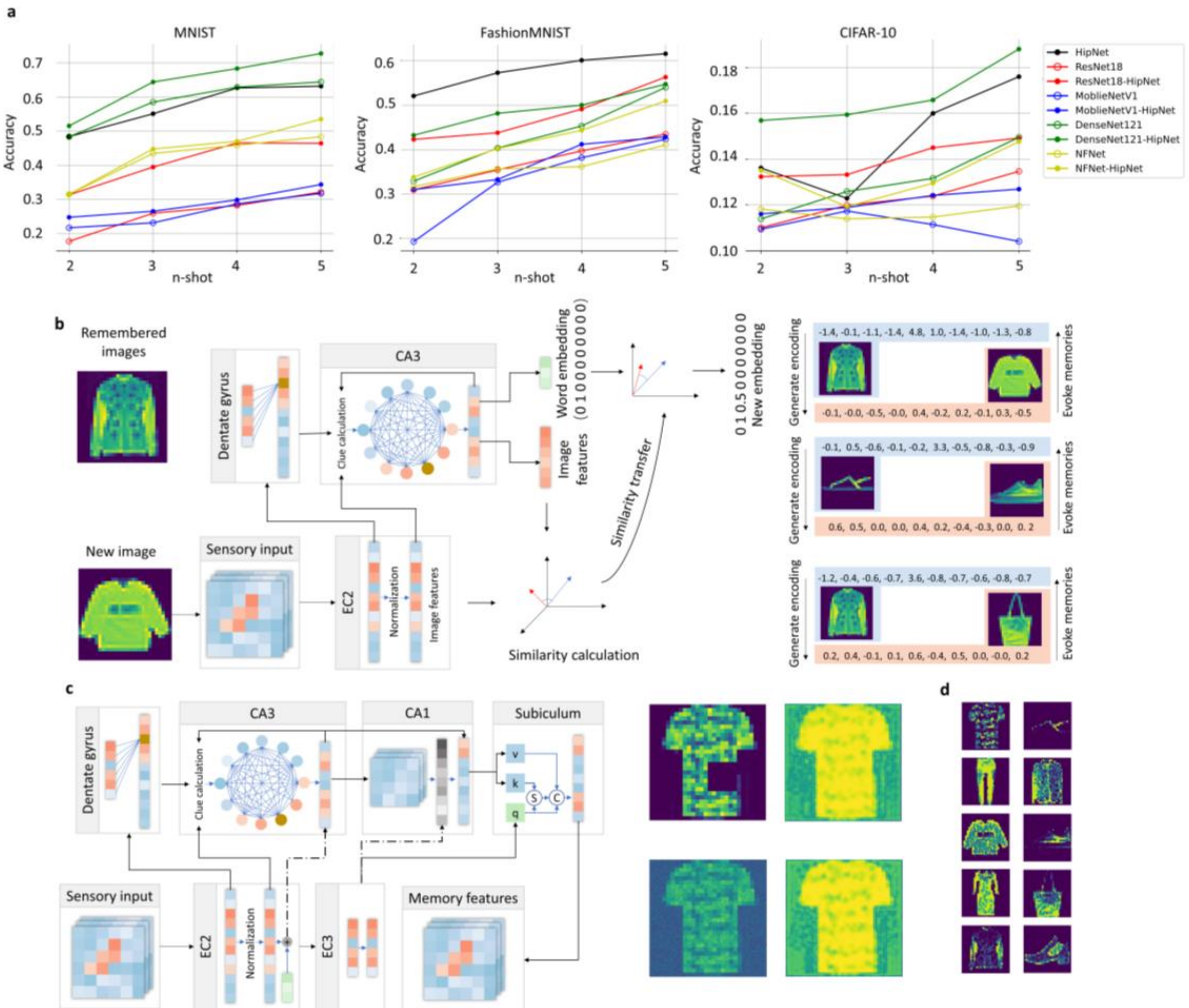

**Fig. 2 | Results of multiple experiments. a,** Comparative results on few-shot image classification. **b,** Comparative results on noisy few-shot image classification. **c,** Left: the model architecture used for constructing word embeddings. Right: visualization of the constructed word embeddings, where memorized word embeddings are shown with a blue background and generated word embeddings are shown with an orange background. **d,** Left: the model architecture used for image restoration. Right: results of image completion and image denoising. **e,** Image generation results.

The experimental results are shown in Fig. 2a. We compare the performance of mainstream convolutional neural network models in recent years [63-66], as well as their performance after incorporating the HipNet structure. We also evaluate the performance of HipNet using a dedicated feature extractor (see the Supplementary Materials). The tests are conducted on 2-shot to 5-shot image classification tasks. To avoid the randomness caused by limited samples, the final accuracy of all results is averaged over 10 independent runs, including this experiment and the subsequent ablation experiments. It can be observed that, across all three datasets, different image classification models achieve improved performance after incorporating the HipNet structure. HipNet with the dedicated feature extractor achieves the best performance on the FashionMNIST dataset, while its performance on the other datasets is second only to ResNet18 equipped with HipNet. For datasets with large variations in object poses, such as CIFAR-10, too few training samples lead to poor test performance. Nevertheless, incorporating the HipNet structure still improves the performance of the corresponding models, which may indicate that the model learns deeper associations between memorized features and current image features, rather than merely comparing image similarity.

**5.4 Performance on Multimodal Tasks**

**Constructing word vectors**

The visual features of different targets have an important correlation with their corresponding semantic similarity, which plays a crucial role in constructing word vectors. By leveraging the memory capabilities of HipNet, it is convenient to calculate the similarity between memorized images and current (new category) images. Subsequently, this image feature similarity, along with the baseline memory's textual word vectors, can be used to automatically infer the word vectors of new categories that align with the image similarity relationship. As shown in Fig.2b, when a new category image is input into HipNet to obtain a new word vector based on previous memory, the cosine similarity between the memory image features output by CA3 and the current image features is first calculated. The method involves performing average pooling on the corresponding channel features of both, then calculating the cosine similarity for each, and finally computing the mean of all channel cosine similarities. This mean is used as the cosine similarity $S_{\text{fea}}$ between the memory image and the current image features. The method to obtain the word vector $\boldsymbol{b} \in \mathbb{R}^{1\times n}$ for the current image using the memory image's corresponding word vector $\boldsymbol{a} \in \mathbb{R}^{1\times n}$ and cosine similarity $S_{\text{fea}}$ is as follows:

First, construct a vector $\boldsymbol{b_0} \in \mathbb{R}^{1\times n}$ that is orthogonal to $\frac{\boldsymbol{a}}{\|\boldsymbol{a}\|_2}$:

$$\boldsymbol{b_0} = \boldsymbol{c} - \boldsymbol{c} \cdot \frac{\boldsymbol{a}}{\|\boldsymbol{a}\|_2} \circ \frac{\boldsymbol{a}}{\|\boldsymbol{a}\|_2} \tag{1}$$

where $\boldsymbol{c} \in \mathbb{R}^{1\times n}$ is a random vector, $\cdot$ represents the dot product, and $\circ$ represents the Hadamard product. It is easy to verify that $\boldsymbol{b_0}$ is orthogonal to $\frac{\boldsymbol{a}}{\|\boldsymbol{a}\|_2}$, i.e., $\boldsymbol{b_0} \cdot \frac{\boldsymbol{a}}{\|\boldsymbol{a}\|_2} = \frac{\boldsymbol{a}}{\|\boldsymbol{a}\|_2}^{\mathrm{T}} \boldsymbol{b_0} = 0$.

Then, using $\frac{\boldsymbol{a}}{\|\boldsymbol{a}\|_2}$ and $\boldsymbol{b_0}$, we can construct a vector $\boldsymbol{b}$ that satisfies the cosine similarity requirement:

$$\boldsymbol{b} = S_{\text{fea}} \cdot \frac{\boldsymbol{a}}{\|\boldsymbol{a}\|_2} + \sqrt{1 - S_{\text{fea}}^2} \cdot \frac{\boldsymbol{b_0}}{\|\boldsymbol{b_0}\|_2} \tag{2}$$

Clearly, $\frac{\boldsymbol{a}\cdot\boldsymbol{b}}{\|\boldsymbol{a}\|_2\|\boldsymbol{b}\|_2} = \boldsymbol{a} \cdot \boldsymbol{b} = S_{\text{fea}}$.

For simplicity, we use the categories Coat, Trouser, and Sandal to train the sensory input section, DG, and CA3 as the memorized known categories, with their one-hot encoded labels serving as their corresponding word vectors. Then, samples from the categories Pullover, Sneaker, and Bag are used to obtain new word vectors. As shown in Fig.2b, the input of new categories activates the memory feature output and obtains their corresponding word vectors that align with the image feature similarity. Since this method uses image feature similarity as a basis, the construction of word vectors is more interpretable. Additionally, because the most similar memory is automatically activated by the current image as the source for constructing new word vectors, there is no need to traverse many other word vectors, making this method computationally efficient.

**Image generation**

HipNet can also generate images from input word vectors. Here, we treat the one-hot encoded labels as textual word vectors, which are then used as clue features for CA3. After retraining the CA3 module, we feed its output image features into a pre-trained deconvolutional network designed to convert the output features of the sensory input section back into original images, thereby generating an image from memory. The image generation results are shown in Fig.2d. It is noteworthy that this text-to-image method does not employ an end-to-end training approach from text to image. Instead, it uses a two-step process: first obtaining intermediate features from text, and then deconvolving these intermediate features into images. This decouples semantic (intermediate feature) generation from final image generation, providing greater flexibility and interpretability for text-to-image tasks.

**Image completion and image super-resolution**

Memory capabilities of HipNet can be utilized for image restoration. For occluded images, similar images can be recalled to complete the missing parts. Similarly, for blurred images, a high-resolution image can be imagined based on similar images in memory. While the input is disturbed, unlike image classification, the recall of occluded (or blurred) input parts should be dominated by the memory features output by CA3, rather than the current sensory features. This shift in dominance can originate from decisions made by other brain regions, a principle beyond the scope of this study. Here, we only modify part of the logic of the subiculum for this task and demonstrate the image restoration effect. As shown in Fig.2c, when inputting missing or noisy images into a trained HipNet, the model can output completed or higher signal-to-noise ratio images. The advantage of HipNet in image restoration lies in its ability to use memory features to obtain the restored intermediate features, eliminating the need to separately design and run this functional module in complex tasks involving image restoration. This enhances the model's computational efficiency in complex tasks and provides a certain level of interpretability.

**5.5 Ablation Experiments**

For simplicity, we conduct ablation experiments on the FashionMNIST few-shot image classification task. In addition to the three main functional modules of the memory component, namely DG, CA3, and CA1, we also ablate the dedicated feature extractor mentioned above to examine the effectiveness of their designs.

**Input Feature Extractor for Few-Shot Image Classification**. We compare the dedicated feature extractor using the regional perception strategy with a feature extractor that directly applies convolution to the whole image, in order to verify our assumptions about its properties. As shown in Fig. 3a, global convolution tends to extract local edge features while retaining edges from other regions, resulting in high similarity among multiple output channels. In contrast, the channels produced by regional convolution tend to output local edge or shape features, which increases feature differences across channels and thus facilitates pattern separation in the subsequent DG module and pattern completion in the CA3 module. The histogram of DG winning-neuron positions and counts shown in Fig. 3b illustrates the effects of the output features generated by the two feature extraction methods on the DG. It can be observed that regional convolution leads to a more uniform distribution of DG winning neurons. By contrast, global convolution causes greater similarity among channel features, resulting in a more concentrated distribution of DG winning-neuron positions. The effects on CA3 are shown in Fig. 3c. The output of CA3 using regional convolution features is closer to the feature distribution output by the convolutional component, and these memory features are more similar for objects with similar shapes, making the memory more intuitively interpretable. In contrast, the CA3 output using global convolution features is more random, making it difficult to explain its relationship with perceptual input features. This may indicate that global convolution is unfavorable for CA3 convergence. Fig. 3d shows the effects of regional and global convolution on few-shot classification performance. When fewer samples are available, regional convolution performs better, reflecting that memory has a stronger influence on HipNet learning in few-shot settings. As the number of samples increases, HipNet learning gradually becomes more dependent on the fitting ability of the task output component.

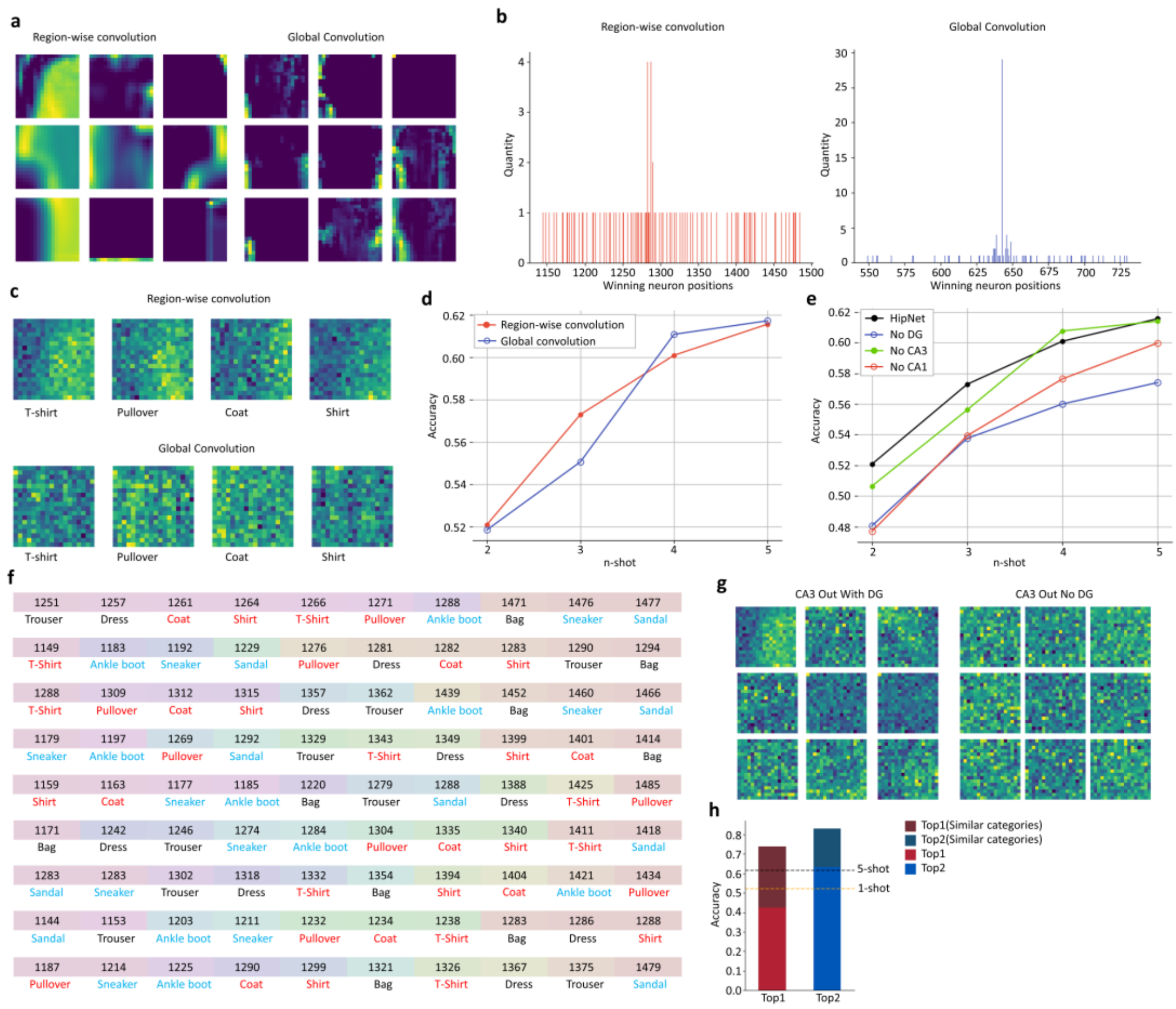

**Fig.3| Analysis of the properties of each HipNet module. a,** Output features of sub-regional convolution and global convolution. **b**, Impact of sub-regional convolution and global convolution on the distribution of DG winning neuron positions. **c**, Effects on CA3 output caused by region-wise convolution and global convolution. **d**, Impact of region-wise convolution and global convolution on final classification accuracy. **e**, Impact of DG, CA3, and CA1 ablation on final classification accuracy. **f**, DG winning neuron positions for different channels and categories, with each row representing a channel. **g**, Impact of DG ablation on CA3 output features. **h**, Top-1 and Top-2 accuracy of CA3 output.

**DG.** To investigate the effect of removing the DG module on HipNet, we directly use the output of EC2 as the cue features for CA3 during training and testing. Fig. 3g shows the output features of CA3 after removing the DG module. It can be observed that, without DG, the features of different channels output by CA3 become more similar to random distributions. This may indicate that removing DG makes the features of different channels more difficult to distinguish, which is analogous to the reduced efficiency of spatial pattern separation caused by DG lesions in biology [67]. Meanwhile, as shown in Fig. 3e, this change leads to a decline in the final performance of the model, further demonstrating the key role of the DG module in HipNet's memory function.

**CA3.** Removing the CA3 module essentially eliminates the memory capability of the model, leaving only the fitting ability of the convolutional neural network composed of the perceptual input component, EC, and task output component. As shown in Fig. 3e, the accuracy degradation caused by removing the memory function is particularly evident when the number of shots is 2 or 3. This again highlights the importance of memory for improving accuracy in few-shot learning.

**CA1.** The ablation of CA1 is implemented by directly replacing the CA1 output with the CA3 output. As shown in Fig. 3e, CA1 ablation substantially affects the performance of the model on few-shot tasks, indicating that the operation of removing instance-specific features in CA1 helps obtain key memories for distinguishing objects. This is also consistent with the phenomenon that CA1 lesions can cause learning impairments [68].

### 5.6 Interpretability

Since HipNet explicitly designs the function of each layer, it exhibits good internal interpretability. Based on the above few-shot image classification experiments, we focus here on analyzing the interpretability of the two modules that have the greatest influence on memory, namely DG and CA3.

**DG.** We record the positions of the final winning neurons in DG. As shown in Fig. 3f, objects that are more similar in morphology, such as upper-body clothing and shoes, have closer winning-neuron positions. This indicates that, within a certain range of EC2 output features, when DG searches for cue features to provide to CA3, it can at least output features from similar categories, which are more similar in distribution and thus facilitate pattern completion in CA3. Although this may lead to memory confusion among several similar categories, this issue could be addressed by top-down modulation that enlarges the feature differences between similar objects; however, designing a more complex feature extractor is beyond the scope of this paper. Nevertheless, this property helps improve CA3 accuracy at the superordinate category level, such as upper-body clothing and shoes. By further determining the specific category based on current sensory features, the model can achieve higher classification performance with limited training samples.

**CA3.** In addition to memorizing image features, CA3 also memorizes the one-hot encoding of the category label corresponding to the current image, similar to multimodal associative memory. We record the memory encoding of category labels output by CA3 during testing and calculate its Top-1 and Top-2 accuracies. Since memory can be easily confused among similar categories, we also record the Top-1 and Top-2 accuracies after grouping morphologically similar categories into broader categories, including upper-body clothing, shoes, dresses, trousers, and bags. As shown in Fig. 3h, the Top-1 and Top-2 accuracies of CA3 are 42.5% and 63.2%, respectively. For the broader categories, CA3 achieves Top-1 and Top-2 accuracies of 73.9% and 83.0%, respectively. This indicates, on the one hand, that CA3 indeed exhibits memory confusion among similar categories and, on the other hand, that CA3 can generate relatively accurate memories at the broader category level. Because the Top-1 accuracy of CA3 is not sufficiently satisfactory, in addition to directly using the CA3 output, we further use the category with the second-highest probability in the CA3 output to reactivate CA3 for a second output. Both outputs are used to train the task output component. This training strategy can explain why the results of few-shot classification from 1-shot to 5-shot gradually approach the CA3 Top-2 accuracy of 63.2%. It is reasonable to infer that, if the memory accuracy of CA3 can be improved, the classification accuracy of HipNet could be further enhanced.

## 6 Conclusion

In this paper, we propose Hippocampus-DETR, a novel memory-enhanced object detection framework that significantly improves the learning efficiency and detection robustness of DETR by introducing a memory module, HipNet, modeled based on the biological structure and function of the hippocampus. To the best of our knowledge, this method is the first to systematically simulate key hippocampal subregions, including EC, DG, CA3, CA1, and the subiculum, from the perspectives of anatomical connectivity and functional differentiation, and to construct corresponding neural network modules, thereby achieving an organic integration of an explicit memory mechanism with object detection. Experimental results show that Hippocampus-DETR achieves leading performance on standard object detection datasets such as MS COCO, while also demonstrating excellent generalization ability and data efficiency across multiple tasks, including few-shot image classification, multimodal feature construction, and image restoration. In addition, the model exhibits good internal interpretability, and the functional effectiveness of each memory submodule is validated through ablation experiments. This study not only advances the application of memory mechanisms in visual tasks, but also provides new insights into the construction of artificial intelligence systems with greater biological plausibility and cognitive capability.

## Supplementary Information

### EC

EC2 performs empty-feature removal and normalization on the features from the sensory input component. Among the features output by the perceptual integration component, empty features whose values are all zero may occur with a very small probability. Such empty features are not suitable for competitive learning in the DG module and also make convergence of the CA3 module more difficult to some extent. Therefore, empty features need to be removed. Specifically, the first element of each empty feature is forcibly replaced with 1. The normalization operation applies min-max normalization to each feature channel to enhance feature stability. Theoretically, EC3 only serves to copy and transmit EC2 information and can be omitted in model implementation. EC5 is a single-layer fully connected network that integrates features from EC3, CA1, and the subiculum. Specifically, these features are concatenated into a one-dimensional vector and then fed into the fully connected layer of EC5.

### DG

The DG module uses a SOM network to perform self-organized learning on the input $\boldsymbol{X}_{\mathbf{EC2}}$ from EC2. First, $\boldsymbol{X}_{\mathbf{EC2}}$ is transformed into a one-dimensional vector and used as the neuronal values of the input layer. Then, the Euclidean distance is used to calculate the similarity between the weight $\boldsymbol{W}_{\boldsymbol{j}}$ of each neuron in the competitive layer and $\boldsymbol{X}_{\mathbf{EC2}}$, and the neuron with the smallest Euclidean distance is identified as the winning neuron. Here, the number of neurons in the competitive layer is set to 1600. Next, the weights of neurons within the effective radius of the topological neighborhood of the winning neuron are updated. The effective radius is calculated as follows:

$$\mathrm{r} = \mathrm{c}_r \cdot e^{-\frac{n_r}{t \cdot log(r_0)}} \tag{3}$$

where r denotes the current effective radius, t=40 is the time constant, $r_0$ is the initial radius and is set to half of the total number of neurons in the competitive layer, $n_r$ is the radius decay coefficient and is equal to the number of SOM training epochs, and $c_r$=40 is the radius adjustment coefficient. For a neuron $j$ within the effective radius of the winning neuron, its weight is updated as follows:

$$\boldsymbol{W_j} = \boldsymbol{W_j} + l \cdot \left(\boldsymbol{X}_{\mathbf{EC2}} - \boldsymbol{W_j}\right) \tag{4}$$

where $\boldsymbol{W_j} \in R^{N_{in}}$ denotes the connection weights between competitive-layer neuron $j$ and all neurons in the input layer, and $N_{in}$ denotes the number of input-layer neurons. Since each token output by the RT-DETR decoder has a length of 256, $N_{in} = 256$. $l$ is the learning rate, calculated as follows:

$$l = l_0 \cdot e^{-\frac{n_l}{t \cdot \log(c_l)}} \tag{5}$$

$$n_l = \frac{n}{d} \tag{6}$$

where $l$ denotes the learning rate used for the current weight update, $l_0$=0.1 is the initial learning rate, $c_l$=5.0 is the learning-rate decay adjustment coefficient, $\mathrm{n} = 5$ is the learning-rate decay coefficient, and $d$ is the Euclidean distance between the weight of the winning neuron and $\boldsymbol{X}_{\mathbf{EC2}}$. After neuron $j$ is updated, max-min normalization is further applied:

$$\boldsymbol{W_j} = \frac{\boldsymbol{W_j} - \min(\boldsymbol{W_j})}{\max(\boldsymbol{W_j}) - \min(\boldsymbol{W_j})} \tag{7}$$

The final output of the DG module is the cosine similarity $\boldsymbol{S} \in \mathrm{R}^{1600}$ between $\boldsymbol{X}_{\mathbf{EC2}}$ and the weight $W_j$ of each neuron in the competitive layer. The DG module is trained for 90 epochs during the training stage.

The process of obtaining the memory cue $\boldsymbol{X}_{\mathbf{clue}}$ from $\boldsymbol{X}_{\mathbf{EC2}}$ and $\boldsymbol{W_j}$ can be viewed as categorizing the current feature into memorized features. Therefore, we treat this process as a fuzzy classification problem. The feature of each memorized channel is used as the kernel of a fuzzy set, and the cosine similarity $\boldsymbol{S}$ is used as the membership degree. Once the current feature $\boldsymbol{X}_{\mathbf{EC2}}$ is known to belong to a memorized feature $\boldsymbol{X}'_{\mathbf{EC2}}$, $\boldsymbol{X}'_{\mathbf{EC2}}$ can be used as the cue feature $\boldsymbol{X}_{\mathbf{clue}}$ for memory activation in subsequent computation. Specifically, during the training stage of the DG module, the 80 winning-neuron positions $\boldsymbol{P}$ and their corresponding weights for the 80 categories in the dataset are saved. Then, only among the positions contained in $\boldsymbol{P}$, the winning-neuron position $\boldsymbol{P_k}$ that maximizes $\boldsymbol{S}$ is selected. Finally, the neuronal weight at position $\boldsymbol{P_k}$ in DG is used as the feature $\boldsymbol{X}'_{\mathbf{EC2}}$ corresponding to the kernel of the fuzzy set.

For the few-shot image classification experiments, the nine channels contained in $\boldsymbol{X}_{\mathbf{EC2}}$ are first separately flattened into one-dimensional vectors $\boldsymbol{X}^{\boldsymbol{i}}_{\mathbf{EC2}}(i = 0,1,\ldots,8)$, which are then used as the neuronal values of the input layer. The Euclidean distance is subsequently used to calculate the similarity between the weight $\boldsymbol{W_j}$ of each neuron in the competitive layer and $\boldsymbol{X}^{\boldsymbol{i}}_{\mathbf{EC2}}$, and the neuron with the smallest Euclidean distance is identified as the winning neuron. Here, the number of input-layer neurons is 324, and the number of competitive-layer neurons is 1800. The calculation of $\boldsymbol{X}_{\mathbf{clue}}$ is as follows. First, during the training stage of the DG module, the final winning-neuron positions $\boldsymbol{P} \in R^{10\times 9}$ and their corresponding weights are saved for 10 categories and 9 channels in the 10-class image classification task, resulting in a total of 90 winning-neuron positions. Repeated positions are allowed. Then, only among the positions contained in P, the winning-neuron positions $\boldsymbol{P_k}(k = 0,1,\ldots,8)$ that maximize $\boldsymbol{S}$ for the corresponding nine channels are selected. Finally, the neuronal weights at positions $\boldsymbol{P_k}$ in DG are used as the features $\boldsymbol{X}_{\mathbf{EC2}}^{i'}(i = 0,1,\ldots,8)$ corresponding to the kernels of the fuzzy sets. As shown in Supplementary Figure 1.

**CA3**

The CA3 module is implemented using the Layer HopfieldPooling module from [51]. The input of this module is the cue feature $\boldsymbol{X}_{\mathbf{clue}}$ from DG and EC2, and its output is the memorized EC2 feature $\boldsymbol{X}_{\mathbf{CA3}}$, including the visual feature $\mathbf{X}_{\mathbf{CA3}}^{\mathbf{v}}$ and $\mathbf{X}_{\mathbf{CA3}}^{\mathbf{other}}$, which is regarded as a feature from another modality. Here, the one-hot encoding of the object category is treated as the memorized encoding of another modality. The loss function is defined as follows:

$$\boldsymbol{L}_{\mathbf{CA3}} = \mathrm{D_{KL}}(\boldsymbol{X}_{\mathbf{CA3}}^{\mathbf{v}} \parallel \boldsymbol{X}_{\mathbf{EC2}}) + \mathrm{MSELoss}(\boldsymbol{X}_{\mathbf{CA3}}^{\mathbf{v}}, \boldsymbol{X}_{\mathbf{EC2}}) + \mathrm{CELoss}(\boldsymbol{X}_{\mathbf{CA3}}^{\mathbf{other}}, \boldsymbol{X}_{\mathbf{other}}) \quad (8)$$

where $\mathrm{D_{KL}}$ denotes the KL divergence, MSELoss denotes the mean squared error loss, $\boldsymbol{X}_{\mathbf{other}}$ is the ground-truth encoding of the other-modality feature, and CELoss denotes the cross-entropy loss. The Adam optimizer is used with a learning rate of 0.00003, and the module is trained for 400 epochs.

For the few-shot image classification experiments, the module outputs the memorized EC2 visual features $\boldsymbol{X}_{\mathbf{CA3}}^{\mathbf{i}}(i = 0,1,\ldots,8)$ and $\boldsymbol{X}_{\mathbf{CA3}}^{\mathbf{other}}$, which is regarded as a feature from another modality. Here, the one-hot encoding of the object category is treated as the memorized target encoding. The loss function is defined as follows:

$$\boldsymbol{L}_{\mathbf{CA3}} = \sum_{i=0}^{8}[\mathrm{D_{KL}}(\boldsymbol{X}_{\mathbf{CA3}}^{i} \parallel \boldsymbol{X}_{\mathbf{EC2}}^{i}) + \mathrm{MSELoss}(\boldsymbol{X}_{\mathbf{CA3}}^{i}, \boldsymbol{X}_{\mathbf{EC2}}^{i})] + \mathrm{CELoss}(\boldsymbol{X}_{\mathbf{CA3}}^{\mathbf{other}}, \boldsymbol{X}_{\mathbf{other}}) \quad (9)$$

where $\boldsymbol{X}_{\mathbf{CA3}}^{i}$ denotes the memory feature of the $i$-th channel output by the CA3 module. The learning rate is set to 0.000001, and the module is trained for 3000 epochs.

**CA1**

The input of the CA1 module is the memory feature $\boldsymbol{X}_{\mathbf{CA3}}$ from the CA3 module. CA1 determines whether this feature is important for the task and outputs the features it considers important as $\boldsymbol{X}_{\mathbf{CA1}}$; otherwise, it outputs empty features. This process is implemented in two steps. In the first step, a convolutional neural network is used to judge the importance of the features in $\boldsymbol{X}_{\mathbf{CA3}}$, outputting the degree of importance of each feature for the current task. This importance is learned by comparing the current output features of EC2 with the output features of CA3 channel by channel. Features that are sufficiently similar can be considered important shared features of the corresponding category; otherwise, they are regarded as unimportant features. Specifically, samples different from those memorized by CA3 are used as inputs. If the cosine similarity between $\boldsymbol{X}_{\mathbf{CA3}}$ and $\boldsymbol{X}_{\mathbf{EC2}}$ is higher than 0.5, the feature of channel $i$ for the current category is considered relatively consistent across different samples and is therefore treated as a shared feature of that category. This feature will be retained in the second step. Conversely, the feature in that channel for that category is considered irrelevant to classification and will be replaced with an empty feature in the second step. The target value, or importance score, of features to be retained is set to 1, whereas that of features to be replaced with empty features is set to 0. MSELoss is used to train the CNN to output these target values. The Adam optimizer is used with a learning rate of 0.0001, and the module is trained for 80 epochs.

The second step converts the importance scores output by the CNN back into features. For the importance score $I$ output by the CNN, if it is greater than or equal to a threshold $\theta$, the corresponding memory feature is considered sufficiently important and the output feature of that channel from CA3 is retained; otherwise, the feature of that channel in CA1 is set to zero:

$$\boldsymbol{X}_{\mathbf{CA1}} = \begin{cases} \boldsymbol{X}_{\mathbf{CA3}}, & I \geq \theta \\ \boldsymbol{O}, & I < \theta \end{cases} \tag{9}$$

where $\boldsymbol{O}$ denotes the zero matrix, and $\theta = 0.7$ is the importance threshold.

For the few-shot image classification experiments, as shown in Supplementary Fig. 2, the first step is modified as follows. A convolutional neural network is used to evaluate the nine channels of each category and outputs whether the feature of each channel is important for the current task, namely image classification. The nine neurons in the output layer of the CNN correspond to the importance scores of the nine channels. This importance is obtained by comparing the current EC2 output feature with the CA3 output feature of the same category channel by channel. Features that are sufficiently similar are regarded as important shared features of the corresponding category; otherwise, they are treated as unimportant features. Specifically, images different from those memorized by CA3 are used as inputs. If the cosine similarity between $\mathbf{X}_{\mathbf{CA3}}^{\mathbf{i}}$ and $\boldsymbol{X}_{\mathbf{EC2}}^{\boldsymbol{i}}$ is higher than 0.5, the feature of channel $i$ for the current category is considered relatively consistent across different samples and thus a shared feature of that category. This feature will be retained in the second step. Conversely, the feature of this channel for that category is considered irrelevant to classification and will be replaced with an empty feature in the second step.

The second step converts the importance scores of the nine channels output by the CNN back into the corresponding nine-channel features. For the importance score $I_i(i = 0,1, \dots ,8)$ of channel $i$ output by the CNN, if $I_i$ is greater than or equal to a threshold $\theta$, the memory feature of this channel is considered sufficiently important, and the output feature of this channel from CA3 is retained. Otherwise, the feature of this channel in CA1 is set to zero:

$$\boldsymbol{X}_{\mathbf{CA1}}^{\boldsymbol{i}} = \begin{cases} \boldsymbol{X}_{\mathbf{CA3}}^{\boldsymbol{i}}, & I_i \geq \theta \\ \boldsymbol{O}, & I_i < \theta \end{cases} \tag{10}$$

where $\boldsymbol{X}_{\mathbf{CA1}}^{\boldsymbol{i}}$ denotes the output of channel $i$ in CA1, and $\theta = 0.7$.

**Subiculum**

The subiculum is implemented using a redesigned attention mechanism. As shown in Supplementary Fig. 4, the output features of CA1 and EC3, denoted as $\boldsymbol{X}_{\mathbf{CA1}}$ and $\boldsymbol{X}_{\mathbf{EC3}}$, are average-pooled and used as the key (k) and query (q), respectively. The cosine similarity between k and q is calculated for each channel and used as the attention score $S_{\text{att}}$. If $S_{\text{att}}$ is greater than or equal to the threshold $\mu$, the output of the subiculum $X_{\text{SUB}}$ retains the features from $\boldsymbol{X}_{\mathbf{CA1}}$. Otherwise, the memory is considered to deviate substantially from the current input, and $X_{\text{SUB}}$ uses the features from $\boldsymbol{X}_{\mathbf{EC3}}$, representing a "distorted" memory:

$$\boldsymbol{X}_{\mathbf{SUB}} = \begin{cases} \boldsymbol{X}_{\mathbf{CA1}}, S_{\text{att}} \geq \mu \\ \boldsymbol{X}_{\mathbf{EC3}}, S_{\text{att}} < \mu \end{cases} \tag{11}$$

where $\mu = 0.6$ is the attention-score threshold.

For the few-shot image classification experiments, as shown in Supplementary Fig. 3, the above operation is adjusted as follows. The output features of CA1 and EC3, denoted as $\boldsymbol{X}^{\boldsymbol{i}}_{\mathbf{CA1}}$ and $\boldsymbol{X}^{\boldsymbol{i}}_{\mathbf{EC3}}$, are average-pooled and used as the key (k) and query (q), respectively. The cosine similarity between k and q is calculated for each channel and used as the attention score $S^{i}_{att}$. If $S^{i}_{att}$ is greater than or equal to the threshold $\mu$, the output of subiculum channel $i$, $\boldsymbol{X}^{\boldsymbol{i}}_{\mathbf{SUB}}$, retains the feature from $\boldsymbol{X}^{\boldsymbol{i}}_{\mathbf{CA1}}$. Otherwise, the memory is considered to deviate substantially from the current input, and $\boldsymbol{X}^{\boldsymbol{i}}_{\mathbf{SUB}}$ uses the feature from $\boldsymbol{X}^{\boldsymbol{i}}_{\mathbf{EC3}}$, representing a "distorted" memory:

$$\boldsymbol{X}^{\boldsymbol{i}}_{\mathbf{SUB}} = \begin{cases} \boldsymbol{X}^{\boldsymbol{i}}_{\mathbf{CA1}}, S^{i}_{att} \geq \mu \\ \boldsymbol{X}^{\boldsymbol{i}}_{\mathbf{EC3}}, S^{i}_{att} < \mu \end{cases} \tag{12}$$

where $\mu = 0.6$.

**Feature Extractor for Few-Shot Image Classification**

For the HipNet-specific feature extractor designed for few-shot image classification experiments, it must not only extract features but also make the extracted features as orthogonal as possible. This is because orthogonalized feature inputs facilitate the pattern completion function of the CA3 module in the memory component, namely recalling an entire event from cues [29], and are therefore crucial for implementing memory function. In this implementation, the input image is first divided into nine smaller regions, and independent convolutional operations are then performed on each region to extract features, which are subsequently fed into the memory component. Compared with applying convolution to the whole image, regional convolution forces the input sources corresponding to the final output channels to be differentiated, thereby facilitating the orthogonalization of output features. Specifically, an input image is first divided into nine smaller regions, and nine groups of convolutional feature extractors are then used to extract features from these regions separately (Supplementary Fig. 4). Each convolutional group ultimately outputs a single-channel feature, due to the computational speed limitation of DG and the convergence performance of CA3. More specifically, the input image size is first standardized to 224 × 224. A sliding window of 112 × 112 with a stride of 56 is used to divide the image into nine partially overlapping subregions. Then, nine groups of convolutional neural networks with mutually independent parameters are used to extract features from each region, with ReLU as the activation function. Each convolutional neural network group finally outputs a one-channel feature. The nine output features are then concatenated along the channel dimension to obtain the output of the sensory input component, $\boldsymbol{X}_{\mathbf{perception}}$. During training, the convolutional neural network is connected to a three-layer fully connected neural network for image classification, with the network parameters shown in Supplementary Fig. 1. Cross-entropy loss and the Adam optimizer are used, with the learning rate set to 0.00001, and the model is trained for 35 epochs.

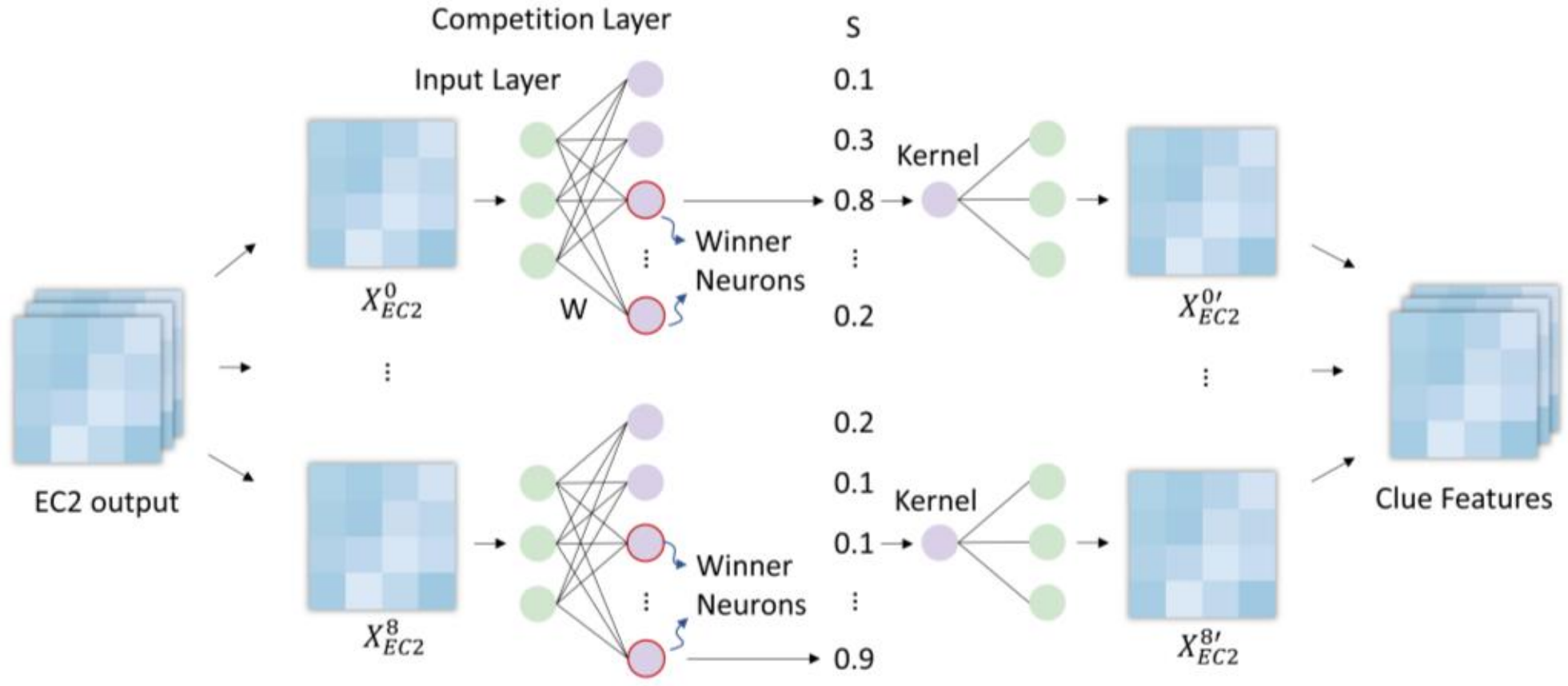

**Supplementary Fig.1| Structure of the DG Module Illustrated by Few-Shot Image Classification.**

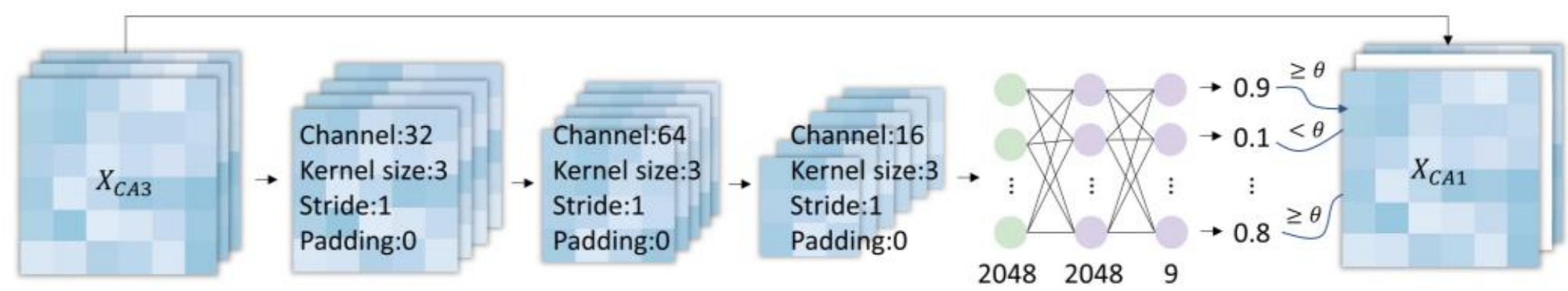

**Supplementary Fig.2| Structure of the CA1 Module Illustrated by Few-Shot Image Classification.**

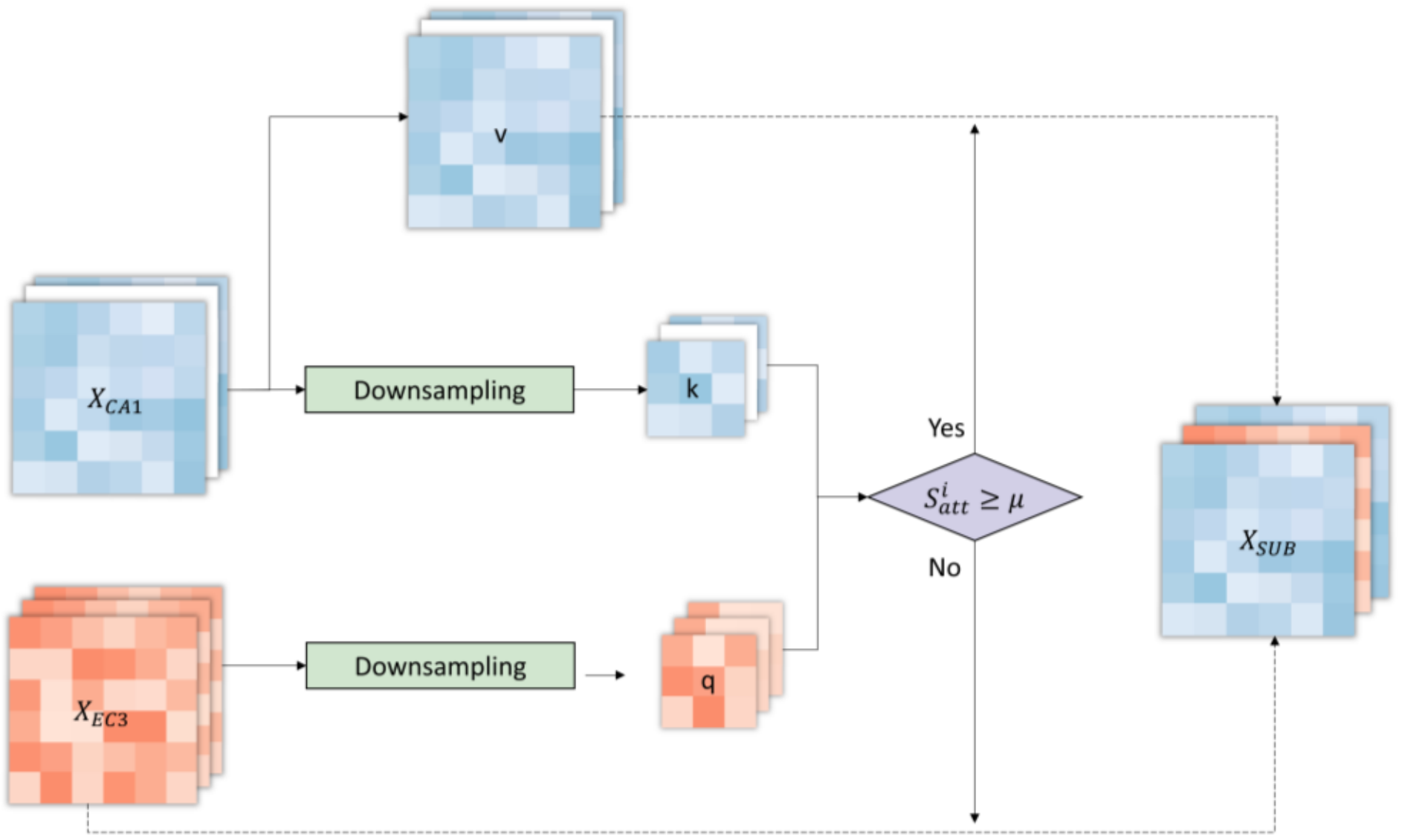

**Supplementary Fig.3| Structure of the Subiculum Module Illustrated by Few-Shot Image Classification.**

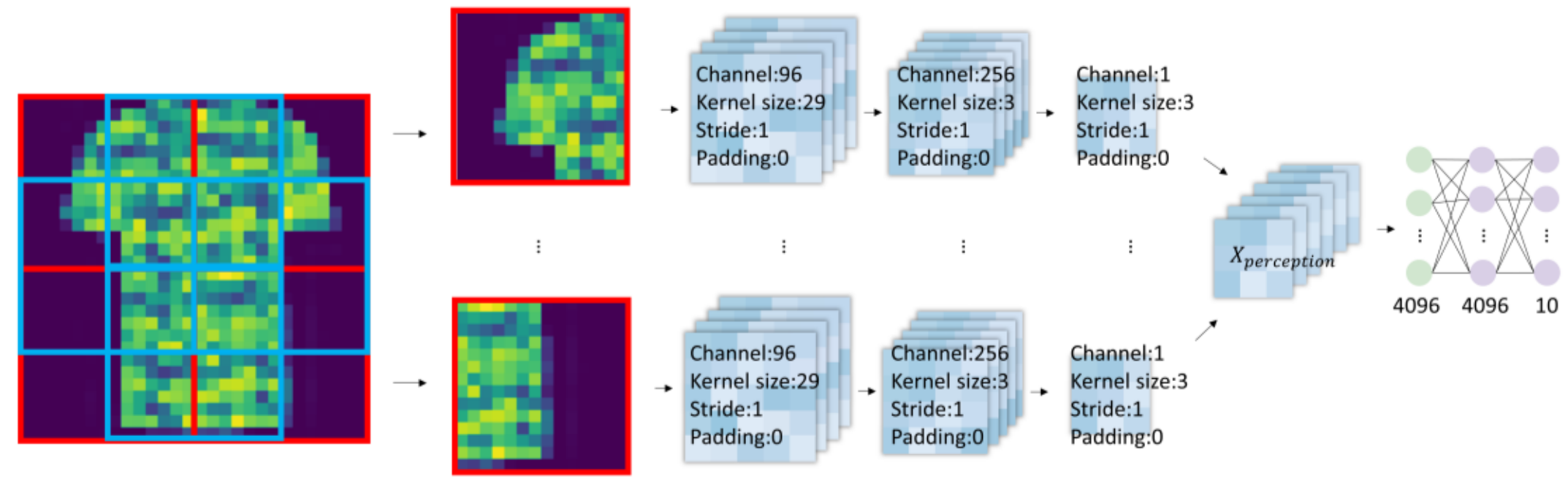

**Supplementary Fig.4| Structure of the Sensory input section Module Illustrated by Few-Shot Image Classification.**